\documentclass{article}
\usepackage{spconf,amsmath,graphicx,hyperref}
\usepackage{booktabs}
\usepackage{siunitx}
\usepackage{pgfplots}
\usepackage[table]{xcolor} 
\definecolor{castblue}{HTML}{1971C2}

\pgfplotsset{compat=1.18}
\newcommand{\CAST}{CAST}

\title{Optimizing Speech Language Models for Acoustic Consistency}
\name{Morteza Rohanian \qquad Michael Krauthammer}

\address{
  University of Zurich, Zurich, Switzerland\\
  \texttt{\{morteza.rohanian, michael.krauthammer\}@uzh.ch}
}
\begin{document}
%
\maketitle
\begin{abstract}
We study speech language models that incorporate semantic initialization and planning losses to achieve robust and consistent generation. Our approach initializes speech tokens with self-supervised features, applies a light alignment loss, and trains with thinning and auxiliary objectives that target robustness and content planning. We train three models: a 0.7B speech-only model, a 1.0B speech-only model, and a 1.0B interleaved model with both text and speech. Acoustic studies show that the speech-only models achieve the highest consistency across speaker, gender, sentiment, room, and background factors, surpassing larger systems. Interleaving improves lexical and syntactic probes and semantic–acoustic alignment but reduces consistency. Linear probes show that our initialization biases the model toward content structure while trading off prosody detail. These results show that LM-side design and training mix control the balance between acoustic stability and semantic grounding without changes to the tokenizer or runtime architecture. A demo and model weights are available for exploration.\footnote{\url{https://mortezaro.github.io/speech-cast/} (demo); \url{https://huggingface.co/KrauthammerLab/cast-0.7b-s2s} (HF model card).}
\end{abstract}

\section{Introduction}
\label{sec:intro}

Speech language models aim to extend the generalization and transfer abilities of text-based language models to spoken inputs and outputs. These models operate directly on discrete speech units and support tasks from semantic understanding to continuation and generation \cite{lakhotia2021generative, borsos2023soundstorm, wang2023neural}. However, robust speech modeling still faces challenges in consistency and invariance. Prior work shows that speech language models struggle with capturing stable speaker identity, background conditions, and prosodic detail across generations, especially under timing variation or noisy inputs. Yet achieving this remains challenging without added complexity.

Most systems introduce architectural changes to improve speech performance. Some add multi-stage decoders or adapter modules, while others incorporate supervised heads or expressive control tokens \cite{defossez2024moshi,nguyen2025spirit}. These methods often increase system complexity and make it difficult to isolate the contribution of the core language model. They blur the line between modeling structure and modeling modality.

\begin{figure}[t]
    \centering
    \includegraphics[width=\columnwidth]{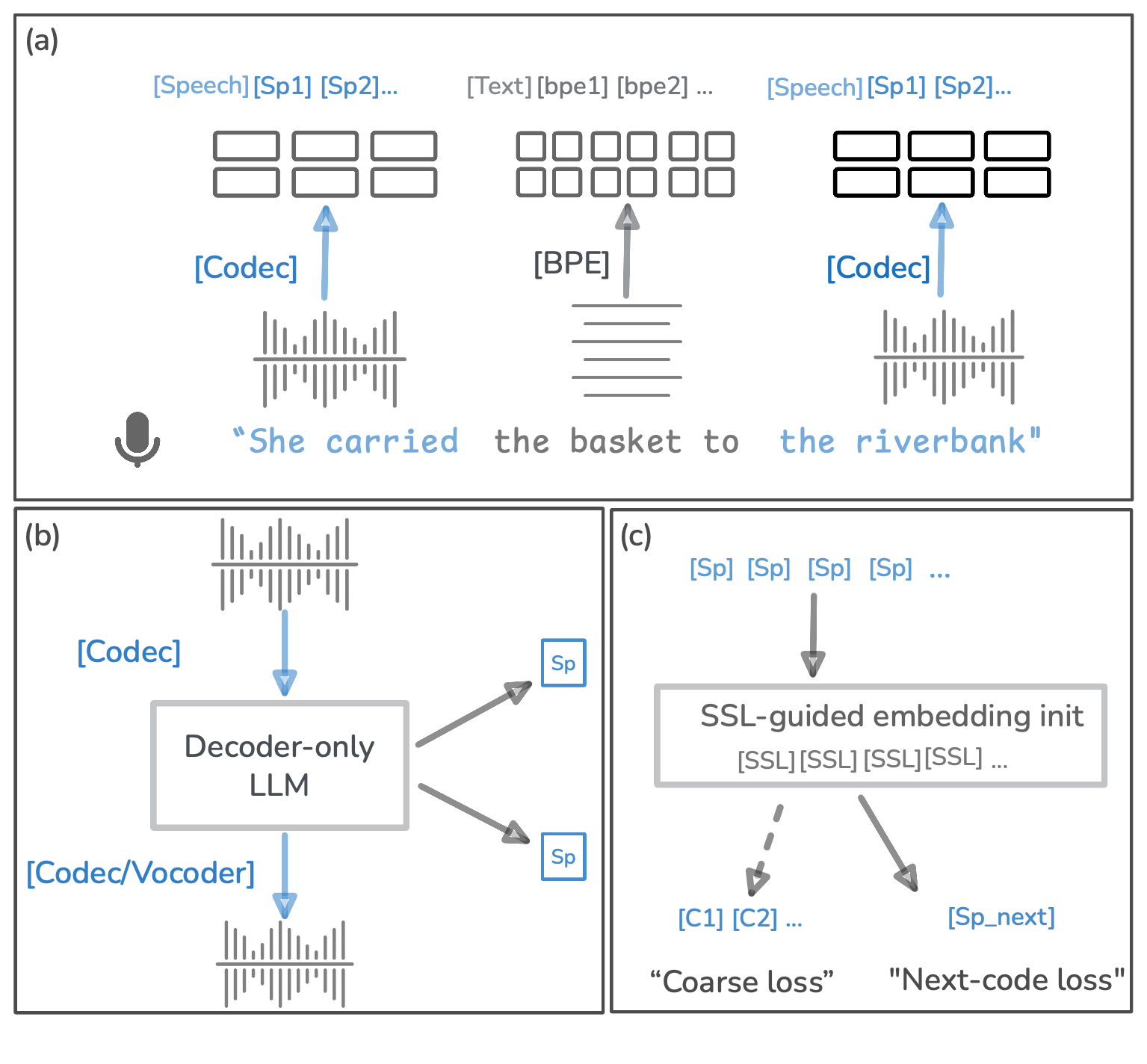}
    \caption{\textbf{Method overview.} 
    (a) Speech (via codec) and text (via BPE) are tokenized and interleaved.
(b) Decoder-only LM predicts unified sequence.
(c) SSL-initialized speech tokens plus auxiliary coarse and next-code objectives improve modeling. \; This design forms the \CAST{} approach.
}
    \label{fig:method}
\end{figure}

Many recent works address speech modeling by exposing codec outputs as discrete tokens in the vocabulary of a decoder-only transformer and by interleaving these tokens with text \cite{nguyen2025spirit}. In this setting, the model predicts a unified sequence of speech and text symbols step by step. Our \CAST{} approach builds on this by shaping the language model’s embeddings and training objectives to prioritize acoustic consistency, isolating LM-side contributions without modifying the tokenizer or inference path.

To support this interface, we shape the language model’s embedding and training strategy. Each speech token maps to a literal token with an embedding initialized from a frozen self-supervised encoder. A stop-gradient alignment loss preserves phonetic structure in the learned representation. During training, we apply consistency-based augmentations: thinning the speech stream at variable rates and erasing short spans to encourage invariance to timing and context gaps. Auxiliary losses guide the model to plan coarse structure before predicting fine acoustic detail.

We train three models: a 0.7B speech-only model, a 1.0B speech-only model, and a 1.0B interleaved model with both text and speech. We evaluate them in three ways. Acoustic studies test whether the model gives higher likelihood to natural recordings compared to samples with mid-utterance changes in gender, speaker, background, room, or sentiment. Semantic studies probe lexical and grammatical regularities in spoken sequences. Alignment studies test whether acoustic conditions match the spoken text. We also include ablations that isolate initialization, thinning, and auxiliary losses.

The 0.7B speech-only model achieves top acoustic consistency (e.g., 90.8\% speaker in SALMON), surpassing larger baselines, indicating stable acoustic planning is a sequence modeling property. Interleaving text and speech reduces consistency but boosts semantic and alignment metrics, suggesting training mix tunes consistency versus grounding without altering tokenizer or architecture.

\section{Related Work}
\label{sec:related}

Revised: Modern speech LMs use neural codecs like SoundStream and EnCodec for high-fidelity, low-bit-rate coding via residual vector quantization. Single-stream tokenizers like WavTokenizer align with decoder-only LMs better than multi-branch designs \cite{ji2024wavtokenizer}. Encoders like HuBERT, WavLM, and Whisper provide semantic tokens or ASR representations for clustering or alignment \cite{hsu2021hubert,chen2022wavlm}. However, codec biases toward reconstruction over invariance motivate our LM-side corrections.

A complementary line of work trains an LM directly on codec (or pseudo-phonetic) units. Early “textless” results show that LMs over self-supervised units capture long-range structure without transcripts \cite{lakhotia2021generative}. AudioLM introduces separate semantic and acoustic token streams to model content and micro-texture \cite{borsos2023audiolm}, while SoundStorm improves efficiency with parallel (non-autoregressive) synthesis for the acoustic stream \cite{borsos2023soundstorm}. VALL-E frames TTS as conditional generation of codec tokens from text, demonstrating strong zero-shot voice transfer \cite{wang2023neural}. Subsequent work scales speech-only LMs on discrete units to strengthen long-context generation \cite{hassid2023textually}.

Recent omni systems extend frozen text LLMs with audio encoders and lightweight adapters to support speech understanding and generation \cite{defossez2024moshi,nguyen2025spirit, nouriborji2025efficient}. These designs broaden task coverage (ASR, captioning, S2S, instruction following) but introduce additional components and training stages that make it harder to isolate what the LM itself contributes. We keep the tokenizer and architecture fixed, adapting only LM-side. This isolates design effects, simplifies deployment, and reveals the stability–grounding trade-off.

\section{Method}

We study a speech LM that treats a frozen neural audio codec as a black box and performs all adaptation on the language model side. The LM acts as a sequencer over discrete codec indices while the frozen codec serves as a vocoder. This separation keeps inference simple and fast and lets us shape audio token embeddings and training signals for robustness invariance and content first planning. The backbone is a decoder-only transformer originally designed for text. Its tokenizer and embedding matrix are extended with speech tokens, so text and speech share a unified vocabulary and prediction objective. We call this LM-side design and training recipe \CAST. Figure~\ref{fig:method} summarizes the design.

\textbf{LM side interface to a frozen codec.}
We use WavTokenizer (single stream SVQ with 4096 codes at 24 kHz) without modifying its encoder or decoder. For each codec index $c\in\{0,\dots,4095\}$ we add a token \texttt{[Sp(c+1)]} to the LM tokenizer and two delimiters \texttt{[Text]} and \texttt{[Speech]} to mark modalities. The BPE core is unchanged. This preserves compatibility with standard text LMs and avoids changes to segmentation or merge rules. 

\textbf{Semantic distilled initialization with alignment.}
Codec codes optimize reconstruction rather than phonetics, so we initialize audio-token embeddings with a speech-aware geometry. Using a frozen SSL encoder HuBERT we compute centroids $\mu_k$ for each code $k$ and project them with a small linear map $P$. We initialize the LM embedding as
\begin{equation}
E_k \leftarrow P(\mu_k) + \varepsilon,\quad \varepsilon\sim\mathcal{N}(0,\sigma^2 I).
\end{equation}
During training we add a light stop gradient alignment
\begin{equation}
\mathcal{L}_{\text{ssl}} = \frac{1}{T_{\text{audio}}}\sum_{t\in\text{audio}}\|h_t - P(\mathrm{SSL}_t)\|_2^2,
\end{equation}
where $h_t$ is the LM hidden state. This initialization lets the LM focus on sequence modeling instead of rebuilding phonetics from scratch.

\begin{table*}[t]
\centering
\caption{SALMON performance (\%). The benchmark reports pairwise preference performance for acoustic consistency (left block) and semantic–acoustic alignment (right block). Best results per column are highlighted.}
\label{tab:salmon}
\small
\setlength{\tabcolsep}{4pt}
\resizebox{\textwidth}{!}{%
\begin{tabular}{l c c c c c c c c}
\toprule
& \multicolumn{6}{c}{\textbf{Acoustic Consistency}} & \multicolumn{2}{c}{\textbf{Semantic–Acoustic Alignment}}\\
\cmidrule(lr){2-7}\cmidrule(lr){8-9}
\textbf{Method} &
{\textbf{Sentiment} $\uparrow$} &
{\textbf{Speaker} $\uparrow$} &
{\textbf{Gender} $\uparrow$} &
{\textbf{Bg (domain)} $\uparrow$} &
{\textbf{Bg (rand.)} $\uparrow$} &
{\textbf{Room} $\uparrow$} &
{\textbf{Sentiment} $\uparrow$} &
{\textbf{Background} $\uparrow$} \\
\midrule
CAST 0.7B (Speech-only)     
& \cellcolor{castblue!20}\textbf{81.8} & \cellcolor{castblue!20}\textbf{90.8} & \cellcolor{castblue!20}\textbf{90.0} & \cellcolor{castblue!20}\textbf{80.0} & \cellcolor{castblue!20}\textbf{77.5} & 90.0 & 51.0 & 56.0 \\
CAST 1B (Speech-only)       
& \cellcolor{castblue!20}\textbf{81.8} & 90.0 & \cellcolor{castblue!20}\textbf{90.0} & 78.0 & 68.5 & \cellcolor{castblue!20}\textbf{91.0} & 48.5 & 51.5 \\
CAST 1B (Speech+Text)       
& 73.0 & 83.5 & 83.5 & 75.0 & 71.5 & 84.5 & 54.5 & 58.0 \\
SpiritLM 7B (Expr.)                  
& 73.5 & 81.0 & 85.0 & 55.0 & 64.0 & 55.5 & 52.0 & \cellcolor{castblue!20}\textbf{59.5} \\
Twist 7B                             
& 61.5 & 71.0 & 70.0 & 55.0 & 60.5 & 62.0 & 51.5 & 54.5 \\
LAST 1.3B                            
& 65.0 & 64.5 & 68.5 & 56.0 & 61.0 & 62.5 & 53.5 & 53.0 \\
Flow-SLM 1B-ext                      
& 65.0 & 76.5 & 80.0 & 70.0 & 64.5 & 73.5 & \cellcolor{castblue!20}\textbf{57.0} & 53.0 \\
\midrule
\textbf{Human}                       
& 97.2 & 91.5 & 98.6 & 83.1 & 88.7 & 94.4 & 93.3 & 95.8 \\
\bottomrule
\end{tabular}%
}
\end{table*}

\textbf{Consistency training.}
We train the LM with two mechanisms that improve robustness invariance. (i) Multi rate thinning and span erasure. For a sampled rate $r\in\{1,2,3,4\}$ we keep indices $\{t: t\bmod r=0\}$ and then erase random spans with probability $p_{\text{erase}}$, producing a sparser sequence. Labels are matched to surviving positions, teaching prediction under timing jitter and missing context. (ii) Delayed coarse and fine auxiliaries. We cluster centroids into $K$ buckets and assign each audio token a coarse label $b_t\in\{1,\dots,K\}$. We add a coarse loss $\mathcal{L}_{\text{coarse}}$ on $b_t$ and a next code loss $\mathcal{L}_{\text{next}}$ on $y_t$, which biases the model to plan content before predicting fine acoustic detail. These augmentations target invariance to timing jitter and robustness under missing context.

\textbf{Interleaved training.}
Beyond speech-only sequences we train on interleaved text–audio streams. For paired audio and transcripts we tokenize text with BPE and audio with codec units and merge them in chronological order using lightweight time alignment. The objective remains next token prediction over the mixed vocabulary. Thinning and erasure are applied only to the audio subsequence, encouraging the LM to anchor acoustic detail in lexical content.

\textbf{Setup} We train on LibriLight English \cite{kahn2020libri} (57k hours), which consists of unlabeled English audiobook recordings (mainly read speech), curated to be relatively clean. We also add the clean subset of People’s Speech \cite{galvez2021people}, contributing about 20k hours of conversational and broadcast data. Unless noted all experiments use the combined dataset.  We train three \CAST{} variants: \CAST{} 0.7B (speech-only), \CAST{} 1.0B (speech-only), and \CAST{} 1.0B (speech+text interleaved).
All models use the Gemma 3 1B decoder-only transformer. We hold the transformer body fixed and vary only the text vocabulary size. The Speech-only 1.0B model uses 262k text tokens plus 4096 speech tokens. The Speech-only 0.77B model uses 56k text tokens plus 4096 speech tokens, reducing parameters by about 24 percent without changing attention. We also train an Interleave 1.0B model with the full text vocabulary and mixed speech text spans. In the interleaved regime each sequence alternates between speech and text spans, with the text covering 35–55 percent of the utterance duration and inserted at random positions. This balances modalities and preserves acoustic continuity. Training uses a constant learning rate of $3.0\times10^{-5}$, an effective batch size of 16 per device, bfloat16 precision, and identical optimizer settings across models.  

\textbf{Inference and scoring.}
At generation logits are masked to $\{\texttt{[Sp*]},\texttt{</s>}\}$ and decoded with the frozen codec. For evaluation we resample inputs to 24 kHz before tokenization and report length normalized negative log likelihood
\begin{equation}
\bar{\ell}=\frac{\sum_{t} m_t\bigl[-\log p(x_t\mid x_{<t})\bigr]}{\sum_t m_t},
\end{equation}
with $m_t$ as a pad mask. The reported score is $-\bar{\ell}$.

\label{sec:typestyle}


\section{Results}

We evaluate the three \CAST{} variants against prior systems that represent the main approaches in the field. For comparison, we include prior systems that represent the main approaches in the field: Twist (7B speech-only) \cite{hassid2023textually}, which initializes from a large text LM and models HuBERT units; LAST (1.3B) \cite{turetzky2024last}, which uses LM-guided tokenization to better align speech tokens with text; SpiritLM (7B interleaved) \cite{nguyen2025spirit}, which extends a text LM with speech tokens and shows strong semantic grounding; and Flow-SLM (1B) \cite{chou2025flow}, which emphasizes efficiency through flow-based objectives. These systems differ in tokenization, architecture, and training objectives, and they provide reference points for evaluating the effects of our LM-side interface and training mix.  We summarize results across SALMON acoustic and alignment tasks \cite {maimon2025salmon} (Table \ref{tab:salmon}), semantic probes (Table \ref{tab:linguistic}), and ARCH linear probes \cite{la2024benchmarking} (Table \ref{tab:sem-distill-stage}).  

\textbf{Acoustic studies.}  
SALMON measures whether the model assigns higher likelihood to a natural recording compared to one where a single acoustic factor (speaker, gender, sentiment, background, or room) changes mid-utterance. For example, gender consistency tests whether a switch from male to female voice inside the same sentence is penalized. The speech-only 0.7B model achieves the strongest stability, scoring 90.8 on speaker consistency and 90.0 on gender consistency, outperforming the 1.0B speech-only model and even larger baselines such as SpiritLM 7B (81.0 speaker, 85.0 gender). This suggests that stability could be more influenced by LM-side training than by parameter count. Interleaving text with speech reduces stability by 5--9 points across all factors, despite using the same codec and decoding path, which ties the drop to the training mix. Across tasks, gender is the easiest factor, while background and room are the most difficult, reflecting that identity cues are localized while scene cues require longer context.  

\begin{table}[t]
\centering
\caption{sWUGGY and sBLiMP performance (\%). Best results per column are highlighted.}
\label{tab:linguistic}
\small
\setlength{\tabcolsep}{6pt}
\begin{tabular}{l c c}
\toprule
\textbf{Method} & \textbf{sWUGGY} & \textbf{sBLiMP} \\
\midrule
CAST 0.7B (Speech-only)   & 65.6 & 55.9 \\
CAST 1B (Speech-only)     & 67.0 & 57.2 \\
CAST 1B (Speech+Text)     & 73.7 & 58.3 \\
SpiritLM 7B               & 75.5 & 58.3 \\
Twist 7B                  & \cellcolor{castblue!20}\textbf{82.8} & 56.2 \\
LAST 1.3B                 & 73.6 & 55.3 \\
Flow-SLM 1B               & 73.2 & \cellcolor{castblue!20}\textbf{60.0} \\
\bottomrule
\end{tabular}
\end{table}

\textbf{Alignment studies.}  
SALMON also tests whether acoustics match the spoken content by pairing a sample where the background or sentiment is consistent with the text against a mismatched version. The interleaved 1.0B model improves sentiment alignment to 54.5 and background alignment to 59.0, compared to 48.5 and 51.5 for the speech-only 1.0B model. Humans score above 93 on both tasks, leaving a gap in joint reasoning over lexical and acoustic context. The trade-off is clear: speech-only favors stability, while interleaving favors alignment.  

\textbf{Semantic probes.}  
We measure lexical and syntactic knowledge with sWUGGY and sBLiMP, where the model must prefer a real word over a pseudoword or a grammatical sentence over an ungrammatical variant. Interleaving speech and text yields a large gain in sWUGGY (73.7 vs.\ 65.6 for the 0.7B speech-only model) and a smaller gain in sBLiMP (58.3 vs.\ 55.9). Larger models such as Twist 7B reach higher sWUGGY (82.8) but do not improve sBLiMP, showing that scaling helps lexical cues more than syntax. This suggests interleaving lets the LM reuse text priors for speech, strengthening lexical regularities.

\textbf{ARCH linear probes.}  
To study internal representations, we train linear classifiers on frozen LM features from six datasets. ESC-50, US8K, and SLURP probe environmental events and intent (content-leaning), while VIVAE and EMOVO probe vocal imitations and emotion (prosody-leaning), and RAVDESS probes acted emotions (mixed). Variant A, which initializes speech-token embeddings from SSL centroids and applies a light alignment loss, outperforms a randomly initialized baseline. It matches or surpasses the final baseline accuracy on content-leaning tasks as early as 10\% of training and reaches +5--6 points on ESC-50 and US8K and +6 points on RAVDESS by the end. It maintains parity on SLURP, while giving up 2--3 points on VIVAE and EMOVO. On average, content tasks rise by +16\% relative while prosody tasks dip by --7\%, matching the intended bias toward content.  

\textbf{Auxiliaries and robustness.}  
We ablate the delayed coarse and next-code auxiliary losses, which encourage the LM to plan content before predicting fine acoustics. Removing them reduces stability by 5--7 points on speaker and gender for both the 0.7B and 1.0B models (e.g., speaker drops from 90.8 to 83.5 in the 0.7B model). Background and room consistency also decline but more moderately, suggesting that auxiliaries contribute most strongly to identity cues. A single outlier appears for room in the 1.0B model, which we attribute to instability in a single run, since all other factors degrade smoothly.

Taken together, these studies show that acoustic stability peaks in the speech-only setting, while interleaving improves alignment and semantics at the cost of consistency. The codec, architecture, and decoding path remain fixed, so the operating point is governed by the training mix and LM-side objectives. This makes the trade-off between stability and lexical grounding controllable without architectural changes.

\begin{table}[t]
\centering
\caption{Classification accuracy (\%) vs.\ training stage for a variant with semantic-distilled initialization. Best results per column are highlighted.}
\label{tab:sem-distill-stage}
\small
\setlength{\tabcolsep}{3.5pt}
\resizebox{\columnwidth}{!}{%
\begin{tabular}{l l
c c c
c c c}
\toprule
& & \multicolumn{3}{c}{\textbf{Audio Events}} & \multicolumn{3}{c}{\textbf{Speech}} \\
\cmidrule(lr){3-5} \cmidrule(lr){6-8}
\textbf{Variant} & \textbf{Stage} &
\textbf{ESC-50} & \textbf{US8K} & \textbf{VIVAE} &
\textbf{RAVDESS} & \textbf{SLURP} & \textbf{EMOVO} \\
\midrule
Raw                         & 100\% & 26.5 & 40.2 & 30.0 & 33.1 &  8.0 & 31.2 \\
Sem-distilled      & 10\%  & 27.9 & 40.8 & 27.7 & 33.5 &  7.9 & 29.8 \\
Sem-distilled      & 40\%  & 30.6 & 43.1 & 26.5 & 34.8 & \cellcolor{castblue!20}\textbf{8.1} & 29.3 \\
Sem-distilled & 100\% & \cellcolor{castblue!20}\textbf{32.6} & \cellcolor{castblue!20}\textbf{45.9} & 27.5 & \cellcolor{castblue!20}\textbf{38.9} & \cellcolor{castblue!20}\textbf{8.1} & 29.3 \\
\bottomrule
\end{tabular}%
}
\end{table}

\begin{table}[t]
\centering
\caption{Effect of auxiliary losses on SALMON performance (\%). Best results per column are highlighted.}
\label{tab:aux}
\small
\setlength{\tabcolsep}{4pt}
\resizebox{\columnwidth}{!}{%
\begin{tabular}{l c c c c c c}
\toprule
\textbf{Method} & {Sent} & {Spkr} & {Gen} & {Bg(In)} & {Bg(R)} & {Room} \\
\midrule
CAST 0.7B (+Aux) & \cellcolor{castblue!20}\textbf{81.8} & \cellcolor{castblue!20}\textbf{90.8} & \cellcolor{castblue!20}\textbf{90.0} & \cellcolor{castblue!20}\textbf{80.0} & \cellcolor{castblue!20}\textbf{77.5} & 90.0 \\
CAST 1B (+Aux)   & \cellcolor{castblue!20}\textbf{81.8} & 90.0 & \cellcolor{castblue!20}\textbf{90.0} & 78.0 & 68.5 & \cellcolor{castblue!20}\textbf{91.0} \\
\midrule
CAST 0.7B (–Aux) & 75.5 & 83.5 & 83.0 & 76.0 & 71.0 & 89.5 \\
CAST 1B (–Aux)   & 75.0 & 83.0 & 82.0 & 75.0 & 71.0 & 90.0 \\
\bottomrule
\end{tabular}%
}
\end{table}

\section{Conclusion}
We study speech language models that improve robustness by distilling self-supervised features and guiding prediction with auxiliary planning. We introduce \CAST{}, with self-supervised initialization, light alignment, and robustness objectives. Across diverse evaluations, the 0.7B speech-only model exhibited top acoustic consistency, while interleaving speech and text enhanced semantics and alignment. Notably, our 0.7B model rivals 7B systems, suggesting LM objectives as a scalable path to consistency. Future work can extend this interface to interactive tasks such as spoken dialogue, assistive speech interfaces, and creative generation where the trade-off between acoustic fidelity and lexical accuracy is critical.

\section{Acknowledgements}
This work was supported as part of the Swiss AI Initiative through a grant from the Swiss National Supercomputing Centre (CSCS) under project ID a02 on Alps.

\bibliographystyle{IEEEbib}
\bibliography{strings,refs}

\appendix
\section{Implementation Details}

\subsection{Environment}
Training used PyTorch 2.4 with Transformers 4.44 and CUDA 12.4. All experiments were run on NVIDIA A100 GPUs (80GB) using mixed-precision (\texttt{bfloat16}) training. We employed gradient checkpointing and fused AdamW optimization.

\subsection{Audio Processing}
Audio inputs were decoded to mono at \SI{16}{kHz} and resampled to \SI{24}{kHz} prior to tokenization. To ensure stability, files with missing or unreadable audio were filtered before loading. All data were normalized per utterance to zero mean and unit variance.

\subsection{Speech Tokenizer}
The speech tokenizer (\texttt{WavTokenizer-large-unify-40token}) was frozen during training. We verified codebook coverage after filtering to avoid degenerate sequences. No fine-tuning or retraining of the codec was performed.

\subsection{Vocabulary and Model Adaptation}
Speech tokens \texttt{[Sp1]}–\texttt{[Sp4096]} were appended to the text LM’s vocabulary. The embedding matrix was resized once before training and padded to the nearest multiple of 8 for efficiency. No merges or special segmentation rules were changed.

\subsection{Training Configuration}
We used a constant learning rate of $4\times10^{-5}$ with cosine decay, batch size 16 per GPU, and 3 training epochs. Checkpoints were saved every 2200 steps with RNG state, tokenizer, and processor to support exact resumption.

\subsection{Reproducibility}
All random seeds were fixed globally (Python, NumPy, PyTorch) and stored with checkpoints. Audio augmentation and sequence thinning use deterministic sampling given a fixed seed. Evaluation metrics were averaged over three seeds.

\subsection{Evaluation Setup}
All evaluation audio was resampled to \SI{24}{kHz}. Likelihoods were computed under a pad mask with length normalization. Confidence intervals correspond to 95\% bootstrap estimates over samples.

\end{document}